\documentclass[runningheads,a4paper]{llncs}
\usepackage{amsmath}
\usepackage{graphicx}
\usepackage{times}
\usepackage{amssymb}
\usepackage{caption}
\usepackage{subfigure}
\usepackage{url}
\urldef{\mailsa}\path|lishiruilishirui@gmail.com,|
\urldef{\mailsb}\path|yilmaz.15@osu.edu,|
\urldef{\mailsc}\path|xiao.157@buckeyemail.osu.edu,|
\urldef{\mailsd}\path|lihua@ict.ac.cn.|

\begin{document}
%
\mainmatter  

\title{4D ISIP: 4D IMPLICIT SURFACE INTEREST POINT DETECTION }

\titlerunning{4D ISIP: 4D IMPLICIT SURFACE INTEREST POINT DETECTION}

%
%
\author{Shirui Li$^{1,2,3}$\and Alper Yilmaz$^{4}$\and Changlin Xiao$^{4}$\and Hua Li$^{1,2,3}$}
\authorrunning{4D ISIP: 4D IMPLICIT SURFACE INTEREST POINT DETECTION}

\institute{$^1$Key Laboratory of Intelligent Information Processing \\ Chinese Academy of Sciences, China.\\
	$^2$Institute of Computing Technology Chinese Academy of Sciences,China. \\ 
	$^3$University of  Chinese Academy of Sciences, China.\\
	$^4$The Ohio State University,United States of America.\\
	\mailsa\\
	\mailsb\\
	\mailsc\\
	\mailsd\\}

%
%

\toctitle{Lecture Notes in Computer Science}
\tocauthor{Authors' Instructions}
\maketitle
\begin{abstract}
In this paper, we proposed a new method to detect 4D spatiotemporal interest point called 4D-ISIP(4 dimension implicit surface interest point). We implicitly represent the 3D scene by 3D volume which has a truncated signed distance function (TSDF) in every voxel. The TSDF represents the distance between the spatial point and object surface which is a kind of  implicit surface representation. The basic idea of 4D-ISIP detection is to detect the points whose local neighborhood has significant variations along both spatial and temporal dimensions. In order to test our 4D-ISIP detection, we built a system to acquire 3D human motion dataset using only one Kinect. Experimental results show that our method can detect 4D-ISIP for different human actions. 
\end{abstract}
\section{Introduction}
\label{sec:intro}

Interest point detection has been a hot topic in computer vision field for a number of years. It's a fundamental research problem in computer vision, which plays a key role in many high-level problems, such as activity recognition, 3D reconstruction, image  retrieval and so on. In this paper we proposed a new method to robustly detect interest point in 4D spatiotemporal space (x,y,z,t) for human action recognition. 

3D spatiotemporal interest point (3D STIP) have been shown to perform well for activity recognition and event recognition \cite{wang2009evaluation,zhu2014evaluating}. It used RGB video to detect interest point, which RGB image is sensitive to color and illumination changes, occlusions, as well as background clutters. With the advent of 3D acquisition equipment, we can easily get the depth information.  Depth data can significantly simplify the task of background subtraction and human detection. It can work well in low light conditions, giving a real 3D measure invariant to surface color and texture, while resolving silhouette pose ambiguities. 

The depth data provided by Kinect, however, is noisy, which may have an impact on interest point detection \cite{zhu2014evaluating}. In order to resolve this problem, we acquire a non-noisy human 3D action representation by fusing the depth data stream into global TSDF volume which can be useful for detecting robust interest points. Then, we introduced a new 4D implicit surface interest point (4D-ISIP) as an extension to 3D-STIP  for motion recognition, especially for human actions recognition.  

\section{Related work}
\label{sec:Related work}
Interest point is usually required to be robust under different image transformations and is typically the local extrema point of some domain.  There are a number of interest point detectors \cite{harris1988combined,shi1994good,lowe2004distinctive,rosten2010faster}  for static images. They are widely used for image matching,image retrieval and image classification. For the activity recognition, the spatiotemporal interest points (STIP) detected from a sequence of images is shown to work effectively. The widely used STIP detectors include  three main  approaches: (1) Laptev
\cite{laptev2005space} detected the spatiotemporal volumes with large variation along spatial and temporal directions in a video sequence. A spatiotemporal second-moment matrix is used to model a video sequence.The interest point locations are determined by computing the local maxima of the response function $H = det(M)-k\cdot trace^{3}(M)$, We will give details about this method in Section 3.2. (2) Dollar
\cite{dollar2005behavior} proposed a cubed detector computing the interest point by the local maxima of the response function R, which is defined as: $R=(I\ast g \ast h_{ev})^{2}+(I\ast g \ast h_{od})^{2}$ where g is the 2D gaussian smoothing kernel, $h_{ev}$ and $h_{od}$ are a quadrature pair of 1D Gabor filters. (3) Willems \cite{willems2008efficient} proposed the Hessian detector, which measures the strength of each interest point using the Hessian matrix. The response function is defined as $S=|det(\Gamma)|$, where $\Gamma$ is the Hessian matrix.

The above methods detect the interest points from RGB images. Compared to depth data RGB images they are sensitive to illumination changes, occlusions, and background clutter. As the development of the depth sensor, many STIP detectors had been extended to the depth data. Xia \cite{xia2013spatio} presented a filtering method to extract STIPs from depth videos called DSTIP. Zhang \cite{zhang20114} extracted STIPs by calculating a response function from both depth and RGB channels. 

All the above methods only use partial view of the human body. Holet \cite{holte2012local} used multi-cameras to construct the 3D human action  and then detected STIPs in every single camera view. Following that, they projected STIPs to 3D space to find 4D(x,y,z,t) spatiotemporal interest points. Cho \cite{cho2015volumetric} proposed a volumetric spatial feature representation (VSFR) to measure the density of 3D point clouds for view-invariant human action recognition from depth sequence images. Kim \cite{kim2014view} extracted  4D spatiotemporal interest points (4D-STIP) in 3D space volume sequence reconstructed from multi-views. They detected interest points having large variations in (x,y,z) space firstly, then they check if those interest points have a significant variation in time axis. Kim \cite{kim2014view} used binary volume to represent the whole 3D human and calculate the partial derivatives. 

In this paper, We only use one Kinect to get non-noisy 3D human motion, which is more practical in real applications. We use implicit surface (TSDF volume) to represent the whole 3D human, which provides a way to robustly calculate the partial derivative in 4 directions (x,y,z,t). We directly calculated the 4D ISIP  in (x,y,z,t) space and choose the points which simultaneously have large variations in different four directions. 
 
Utilizing a single Kinect to accurately recover 3D human action is an active and challenging research topic in recent years. Many methods \cite{weiss2011home,chen2013tensor,bogo2015detailed} get 3D human model based on a trained human template, but those methods can't be used to reconstruct the human body with clothes. Zhang \cite{zhang2014quality}  reconstructed human body with clothes. However all of those methods require the reconstructed human to stay still during acquisition, which is not possible in reality. Newcombe \cite{newcombe2015dynamicfusion} proposed a real-time method to reconstruct dynamic scenes without any prior templates. But it is not capable of long term tracking, because of growing warp field and error accumulation. Guo \cite{guo2015robust} introduced a novel $L_0$ based motion regularizer with an iterative optimization solver, which can robustly reconstruct non-rigid geometries and motions from single view depth input. In this paper, we combine Newcombe \cite{newcombe2015dynamicfusion}  and  Guo \cite{guo2015robust}  to build a system to construct 3D human motion dataset for 4D ISIP detection. 

\section{Method}
\label{sec:METHOD}
Before we introduce 4D-ISIP we introduce how to represent and acquire human action dataset firstly. We adopt the volumetric truncated signed distance function (TSDF)
\cite{curless1996volumetric}  to represent the 3D scene. we combine Newcombe \cite{newcombe2015dynamicfusion}  and  Guo \cite{guo2015robust}  to build a system to construct 3D human motion dataset for 4D ISIP detection. Then, we gave a detail about the 3D spatial temporal interest points below. Lastly, we will introduce how to extended 3D-STIP to 4D-ISIP.

\subsection{Acquisition of 3D human motion dataset}

Upon acquisition of every input depth data, we first estimate the ground plane using RANSAC to segment human body from the ground. This is followed by point neighborhood statistics to filter noise outlier data. 

DynamicFusion \cite{newcombe2015dynamicfusion}  can reconstruct non-rigidly deforming scenes in real-time  by fusing RGBD scans acquired from commodity sensors without any template. It can generate denoised, detailed and complete reconstructions. But it is not capable of long term tracking, because of growing warp field and error accumulation. The method proposed by Guo \cite{guo2015robust}  provides long term tracking  using a human template. We use DynamicFusion to get a complete human body mesh by rotating our body in front of the Kinect as the template. Then we use Guo[20] to track human motion with partial data input. 

\begin{figure}
\centering \includegraphics[width=4.5in]{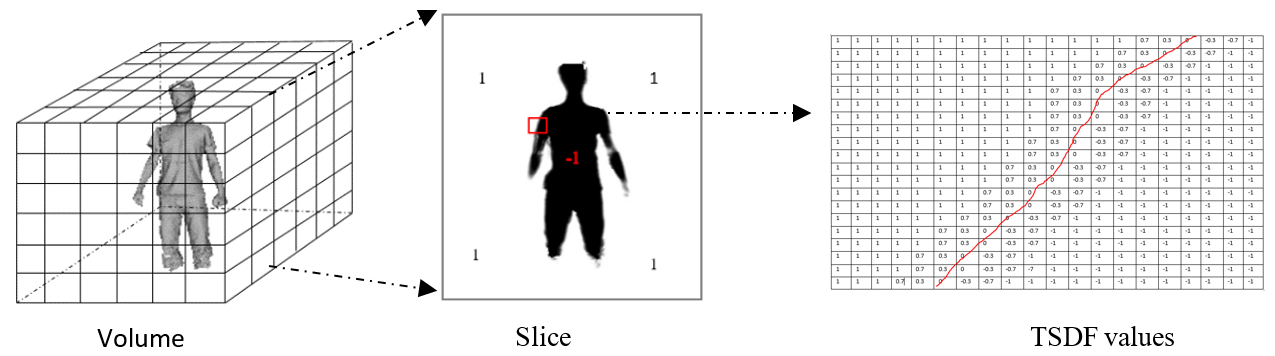} 
\caption{TSDF volume }\label{fig:tsdf} 
\end{figure}

Lastly, We generated the same topology mesh for every motion frame. Then we transformed those mesh into TSDF representation. In  Figure \ref{fig:tsdf}, we illustrate how we use TSDF to implicitly represent an arbitrary surface as zero crossings within the volume. The whole scene is represented by a 3D volume with a TSDF value in each voxel. TSDF is the truncated distance between spatial point and object surface.

\begin{equation}
\Psi(\eta) \left \{ \begin{array}{ll}
min(1,\frac{\eta}{\tau}) & \textrm{ if $\eta \ge -\tau$}\\
-1 & \textrm{otherwise}
\end{array}\right.
\end{equation}
where $\tau$ is the threhold distance and $\eta$ is the distance to surface, $\Psi(\eta)$ is the truncated signed distance. 

\subsection{3D Spatial Temporal Interest Points}
In order to model a spatial-temporal image sequence, Laptev
\cite{laptev2005space} used a function $f:\mathbb{R} ^{2} \times \mathbb{R} \mapsto \mathbb{R}$  and constructed its linear scale-space representation $L:\mathbb{R}^{2} \times \mathbb{R}\times \mathbb{R}_{+}^2\mapsto \mathbb{R}$ by convolution of $f$ with an anisotropic Gaussian kernel with independent spatial variance  $\sigma_{s}^{2}$ and temporal variance ${\sigma_{t}^{2}}$:

\begin{equation}
L(x,y,t; \sigma_{s}^{2},\sigma_{t}^{2}) = g(x,y,t; \sigma_{s}^{2},\sigma_{t}^{2}) \ast f(x,y,t),
\end{equation}

where the spatiotemporal separable Gaussian kernel is defined as:
\begin{equation}
\begin{aligned}
g(x,y,t; \sigma_{s}^{2},\sigma_{t}^{2}) = \frac{1}{\sqrt{(2 \pi)^{3}\sigma_{s}^{4} \sigma_{t}^{2}}} \\
 \times  exp(\frac{-(x^{2}+y^{2})}{2\sigma_{s}^{2}} - \frac{t^{2}}{2 \sigma_{t}^{2}})
 \end {aligned},
\end{equation}

where $\sigma_{s^{'}}^{2} = l \sigma_{s}^{2}$ and $\sigma_{t^{'}}^{2} = l\sigma_{t}^{2}$. Then they define a spatiotemporal second-moment matrix as:
\begin{equation}
M = g(\cdot;\sigma_{s^{'}}^{2},\sigma_{t^{'}}^{2}) \ast 
\left ( \begin {array}{ccc} L_{x}^{2}  & L_{x}L_{y} & L_{x}L_{t} \\
L_{x}L_{y} & L_{y}^{2}  & L_{y}L_{t} \\
L_{x}L_{t}  & L_{y}L_{t} & L_{t}^{2}
\end{array} \right).
\end{equation}

 The first-order derivatives of f are  given by:

\begin{eqnarray}
\begin{split}
L_{x}(\cdot ; \sigma_{s}^{2},\sigma_{t}^{2})= \partial _{x}(g \ast f) \\
L_{y}(\cdot ; \sigma_{s}^{2},\sigma_{t}^{2})= \partial _{y}(g \ast f) \\
L_{t}(\cdot ; \sigma_{s}^{2},\sigma_{t}^{2})= \partial _{t}(g \ast f)
\end{split}.
\end{eqnarray}

For detecting interest points,  search for region in $f$ having significant eigenvalues $\lambda_{1},\lambda_{2},\lambda_{3}$ of $M$. Laptev \cite{laptev2005space}  calculate $H$ by combining the determinant and the trace of $M$. And select the point with a large $H$ value as the STIP:

\begin{equation}
H = det(M) - k\cdot trace^{3}(M) =\lambda_{1} \lambda_{2} \lambda_{3} - k(\lambda_{1}+\lambda_{2}+\lambda_{3})^{3}, 
\end{equation}
where $k = 0.04$ is an empirical value. 

\subsection{4D Implicit Surface Interested Points}

We define $p:\mathbb{R} ^{3} \times \mathbb{R} \mapsto \mathbb{R}$  is a truncated signed distance function which is the distance to it's nearest surface point. This can be regarded as an implicit surface representation. In this paper, our goal is to find interest points that have significant  variation in $(x,y,z,t)$ directions. Firstly, we do a Gaussian filtering for the complete 3D motion sequences. Considering that spatial and temporal directions have different noise and scale charcteristics, we use $\bar{\sigma}_{s}^{2}$ for spatial space scale and $\bar{\sigma}_{t}^{2}$ for temporal scale:
\begin{equation}
\bar{L}(x,y,z,t;\bar{\sigma}_{s}^{2},\bar{\sigma}_{t}^{2}) = \bar{g}(x,y,z,t;\bar{\sigma}_{s}^{2},\bar{\sigma}_{t}^{2}) \ast p(x,y,z,t),
\end{equation}
which results in 4D Gaussian given by:
\begin{equation}
\begin{split}
\bar{g}(x,y,z,t;\bar{\sigma}_{s}^{2},\bar{\sigma}_{t}^{2}) = \frac{1}{\sqrt{(2\pi)^{3}\bar{\sigma}_{s}^{6}\bar{\sigma}_{t}^{2}}} 
\\ \times exp(-\frac{(x^{2}+y^{2}+z^{2})}{2\bar{\sigma}_{s}^{2}}-\frac{t^{2}}{2\bar{\sigma}_{t}^{2}})
\end{split}
\end{equation}
After filtering, we define a spatiotemporal second-moment matrix, which is a 4-by-4 matrix composed of first order of spatial and temporal derivatives averaged by Gaussian weighting function:
\begin{equation}
\bar{M} = \bar{g}(\cdot;\bar{\sigma}_{s^{'}}^{2},\bar{\sigma}_{t^{'}}^{2}) \ast 
\left ( \begin {array}{cccc} \bar{L}_{x}^{2}  & \bar{L}_{x}\bar{L}_{y} & \bar{L}_{x}\bar{L}_{z} & \bar{L}_{x}\bar{L}_{t} \\
                                    \bar{L}_{x}\bar{L}_{y} & \bar{L}_{y}^{2}  & \bar{L}_{y}\bar{L}_{z} & \bar{L}_{y}\bar{L}_{t} \\
                                    \bar{L}_{x}\bar{L}_{z}  & \bar{L}_{y}\bar{L}_{z} & \bar{L}_{z}^{2}& \bar{L}_{z}\bar{L}_{t} \\
                                    \bar{L}_{x}\bar{L}_{t}  & \bar{L}_{y}\bar{L}_{t} & \bar{L}_{z}L_{t}&  \bar{L}_{t}^{2} \\
                                    \end{array} \right),
\end{equation}

where $\bar{\sigma}_{s^{'}}^{2} = l^{'}\bar{\sigma}_{s}^{2}$ and $\bar{\sigma}_{t^{'}}^{2} = l^{'}\bar{\sigma}_{t}^{2}$. $l^{'}$ is an empirical value, in our experiments we set $l^{'}=2$. 

\begin{equation}
\begin{split}
\bar{L}_{x}(\cdot ; \bar{\sigma}_{s}^{2},\bar{\sigma}_{t}^{2})= \partial _{x}(\bar{g} \ast p), \\
\bar{L}_{y}(\cdot ; \bar{\sigma}_{s}^{2},\bar{\sigma}_{t}^{2})= \partial _{y}(\bar{g} \ast p), \\
\bar{L}_{z}(\cdot ; \bar{\sigma}_{s}^{2},\bar{\sigma}_{t}^{2})= \partial _{z}(\bar{g} \ast p),\\
\bar{L}_{t}(\cdot ; \bar{\sigma}_{s}^{2},\bar{\sigma}_{t}^{2})= \partial _{t}(\bar{g} \ast p).
\end{split}
\end{equation}

In order to  extract interest points, we search for regions in $p$ having significant eigenvalues $\bar{\lambda}_{1}<\bar{\lambda}_{2}<\bar{\lambda}_{3}<\bar{\lambda}_{4}$ of $\bar{M}$.  Similar to the Harris corner function and STIP function, we define a function as follows:

\begin{equation}
\begin{split}
\bar{H}=det(\bar{M})-k*trace^{4}(\bar{M}) \\
= \bar{\lambda}_{1}\bar{\lambda}_{2}\bar{\lambda}_{3}\bar{\lambda}_{4}-k(\bar{\lambda}_{1}+\bar{\lambda}_{2}+\bar{\lambda}_{3}+\bar{\lambda}_{4})^{4}.
\end{split}
\end{equation}

 letting the ratios $\alpha = \bar{\lambda}_{2}/\bar{\lambda}_{1}$ ,$\beta = \bar{\lambda}_{3}/\bar{\lambda}_{1}$,$\gamma = \bar{\lambda}_{3}/\bar{\lambda}_{1}$, we re-write $H$ as 
\begin{equation}
\bar{H} = \bar{\lambda}_{1}^{4}(\alpha\beta\gamma-k(1+\alpha+\beta+\gamma)^{4}),
\end{equation}

where $\bar{H}\ge 0$, we have $k\le\alpha\beta\gamma/(1+\alpha+\beta+\gamma)^{4}$. Suppose $\alpha=\beta=\gamma = 23$, we get $k\leq 0.0005$. In our experiment we use $k = 0.0005$. We select point with a $\bar{H}$ value bigger than a threshold value $\bar{H}_{t}$ as the candidate. At last, we select the points with the local maxima $\bar{H}$ as the 4D ISIPs.


\section{Experiments}
\label{sec:typestyle}

\subsection{3D human action reconstruction}

There are a number of datasets for human action recognition. Some of those datasets \cite{rodriguez2008action,niebles2010modeling,tran2008human,kuehne2011hmdb,reddy2013recognizing} are captured by single RGB camera. Some of those datasets \cite{weinland2007action,gkalelis2009i3dpost} are captured by multiple view RGB cameras, which can provide 3D human motion. However, the acquired 3D models are not accuracy enough. There are also some datasets \cite{wang2012mining,li2010action} captured by single Kinect. But those datasets have an only partial view data with high noise. 

\begin{figure} 
	\centering
	\subfigure[Acquisition setup]{ \label{fig:subfig:a}
		\includegraphics[width=0.8in]{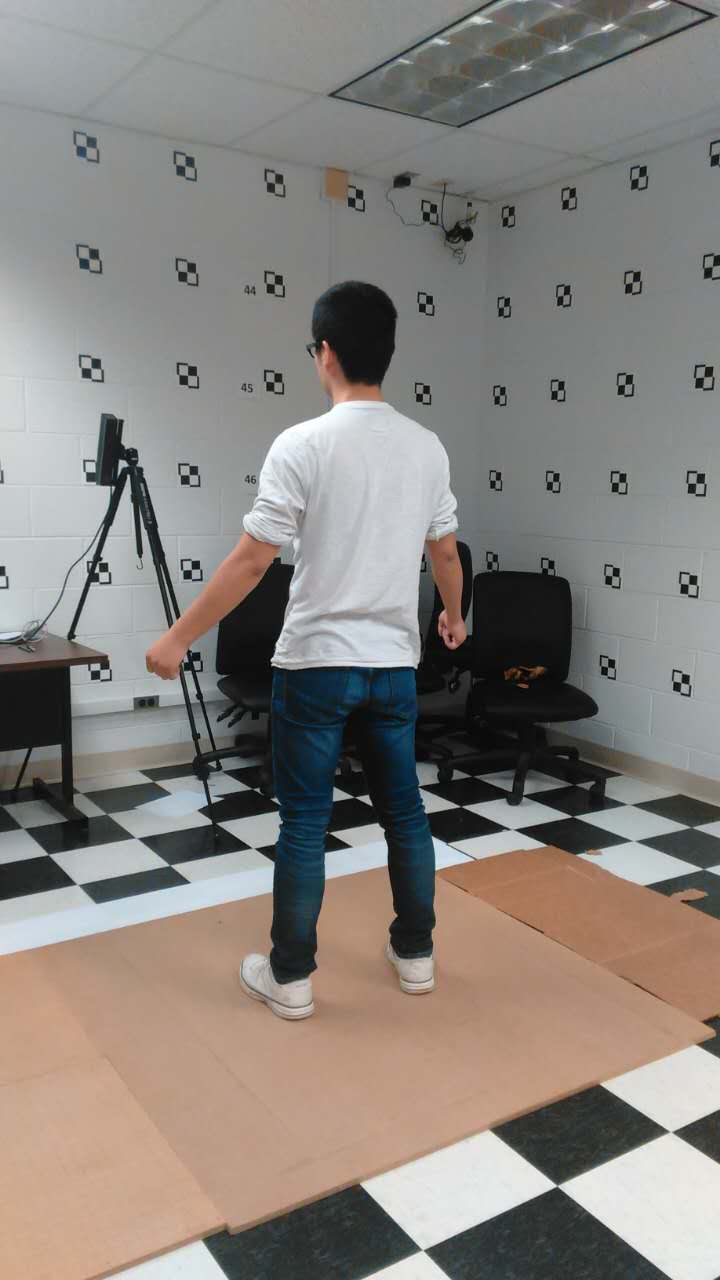}} 
	\subfigure[3D body model]{ \label{fig:subfig:b} 
		\includegraphics[width=1.2in]{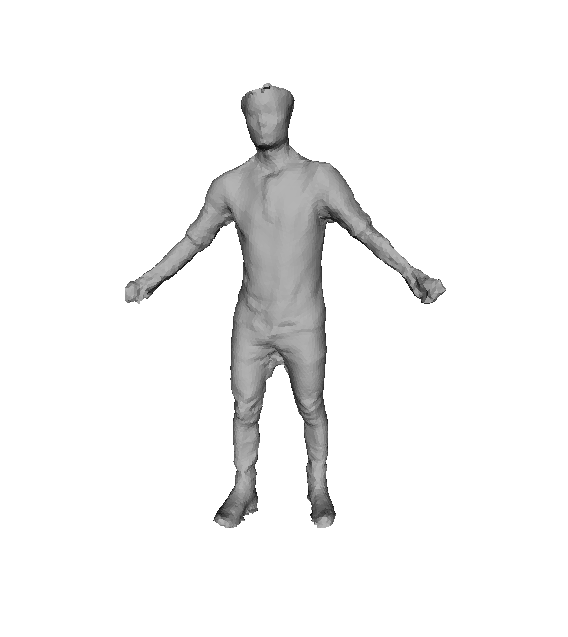}
     	\includegraphics[width=1.2in]{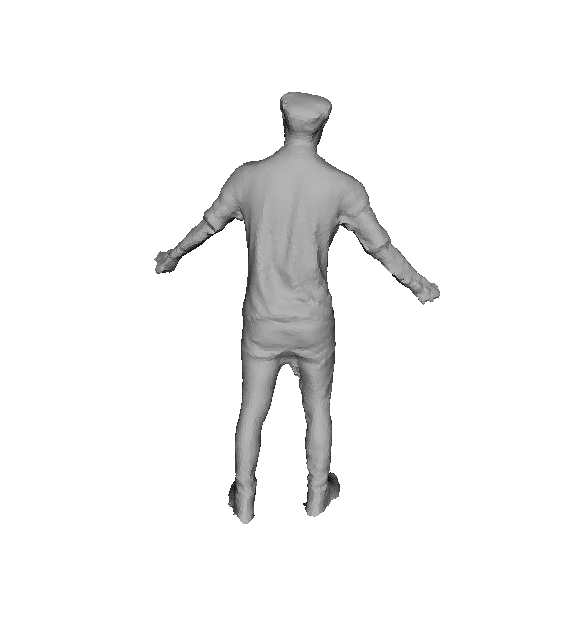}} 
	\caption{Reconstruction setup and results}\label{fig:reconstruction} 
	
\end{figure}

\begin{figure} 
	\centering
	\subfigure[One view input data]{ \label{fig:subfig:a}
		\includegraphics[width=1.5in]{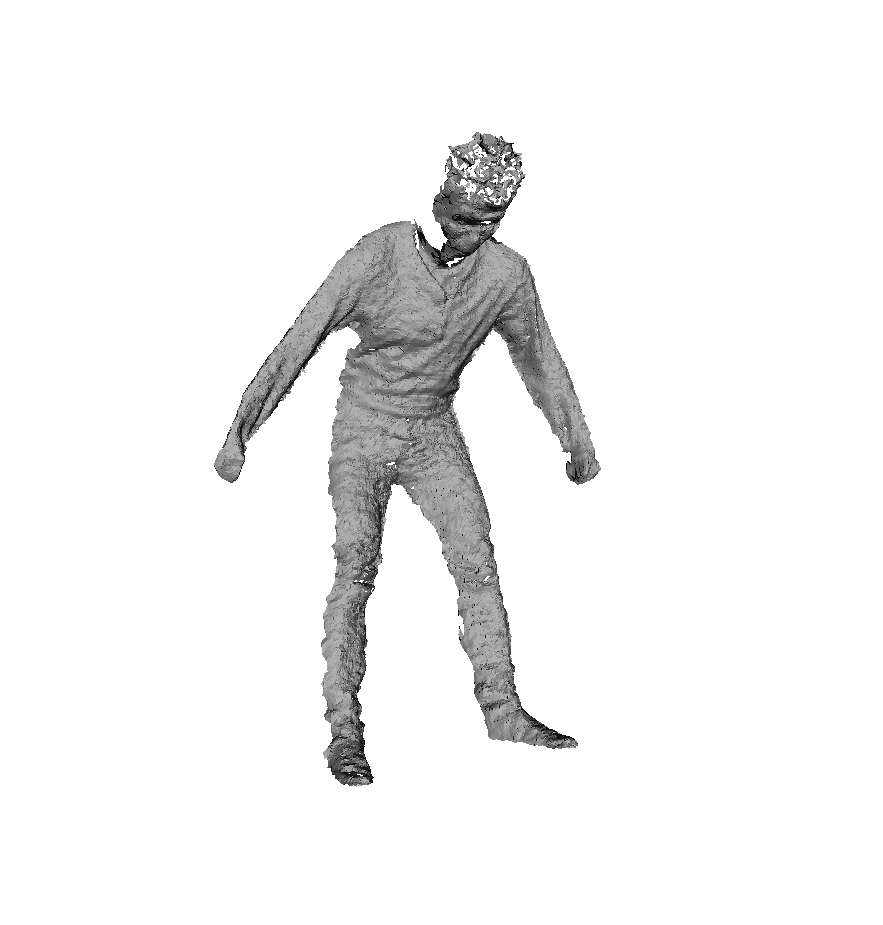}} 
	\subfigure[Complete body model]{ \label{fig:subfig:b} 
		\includegraphics[width=1.5in]{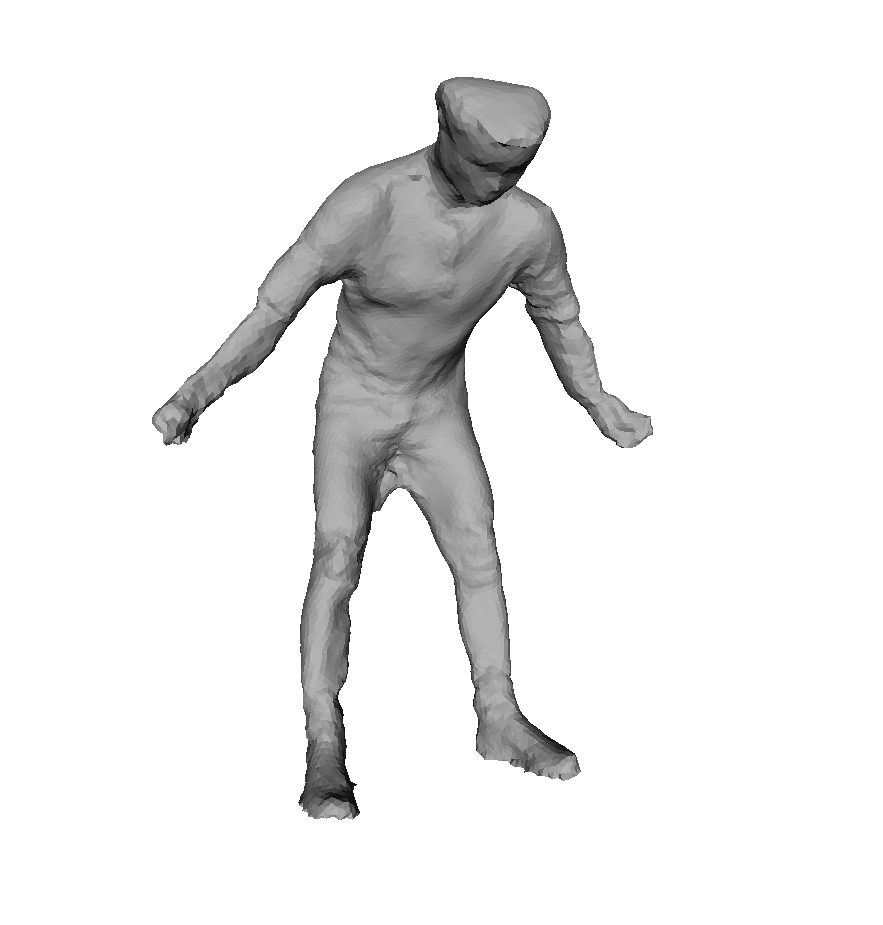}} 
	\caption{Shape completion, The complete 3D body model (b) is driven by the input one view data (a).}\label{fig:shapecompletion} 
\end{figure}

\begin{figure} 
	\centering
	\subfigure[Waving]{ \label{fig:subfig:a} 
		\includegraphics[width=0.8in]{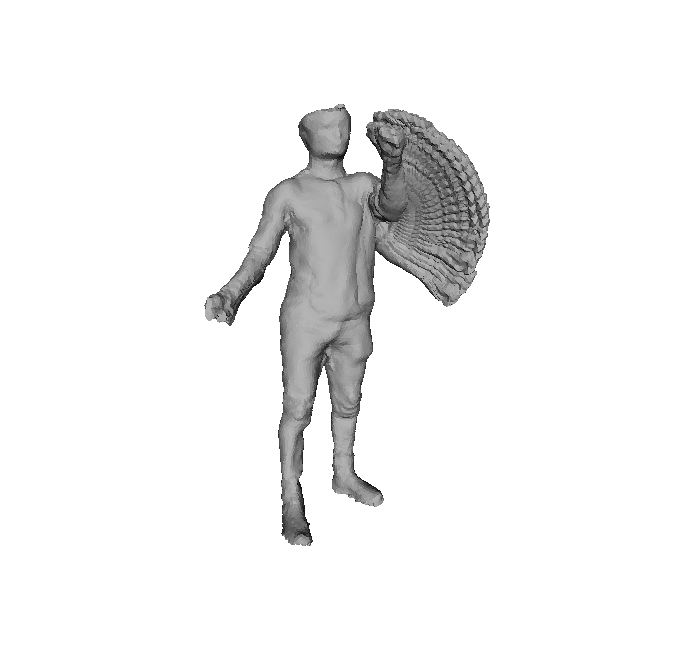}}
	\subfigure[Bowing]{ \label{fig:subfig:b} 
		\includegraphics[width=0.8in]{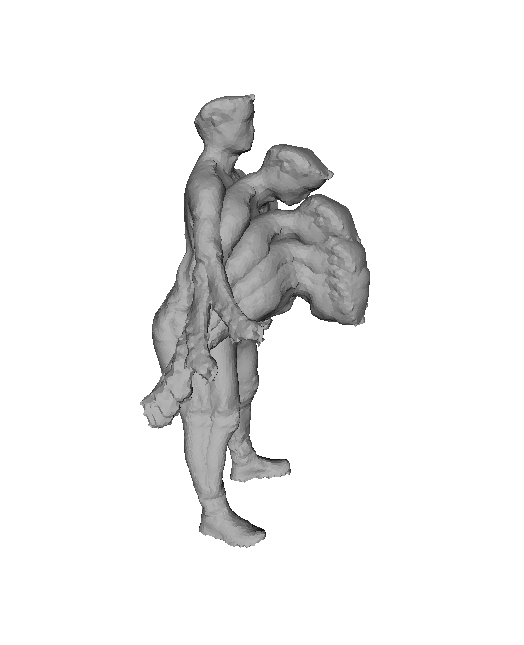}}
	\subfigure[Walking]{ \label{fig:subfig:c} 
		\includegraphics[width=0.8in]{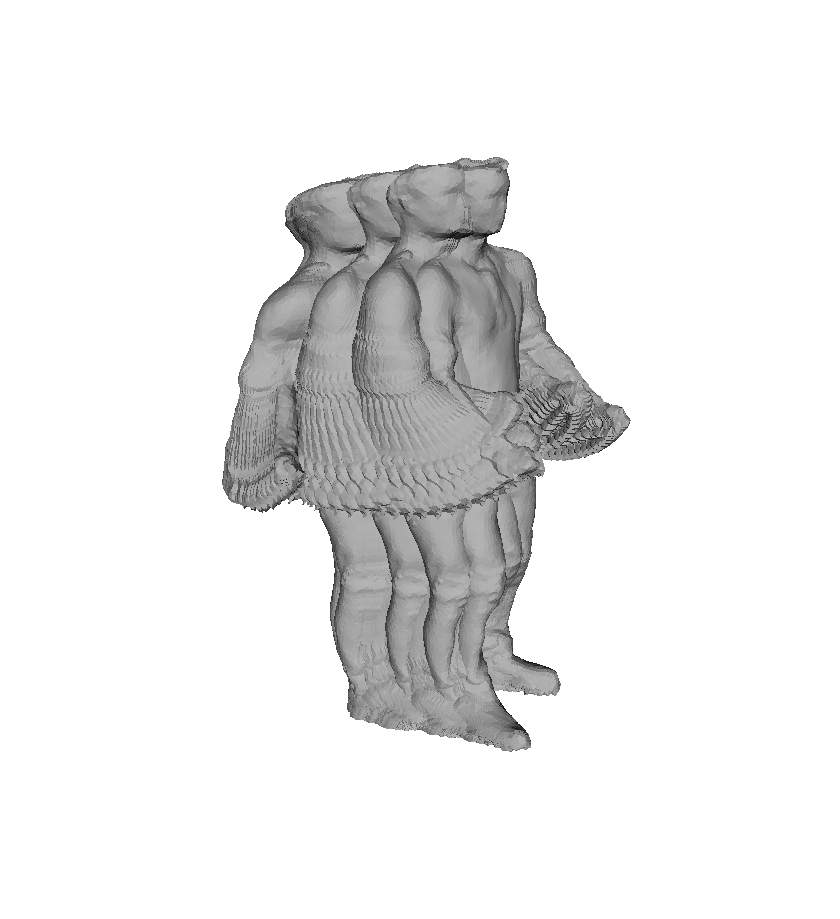}} 
		\subfigure[Kicking]{ \label{fig:subfig:d} 
		\includegraphics[width=0.8in]{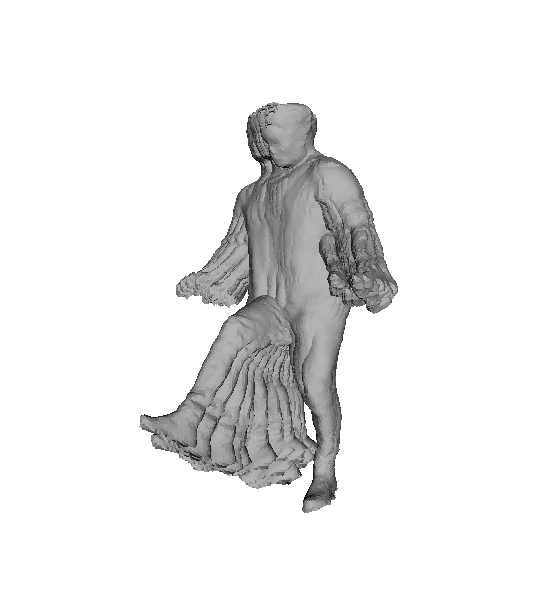}}
	\subfigure[Clapping]{ \label{fig:subfig:e} 
		\includegraphics[width=0.8in]{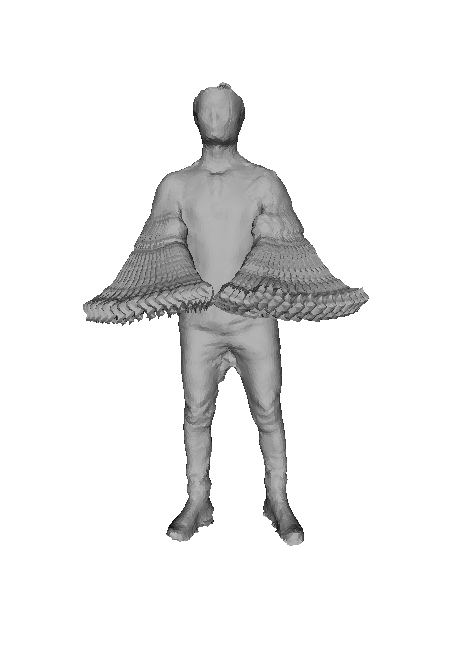}}
	\subfigure[Looking watch]{ \label{fig:subfig:e} 
	\includegraphics[width=0.8in]{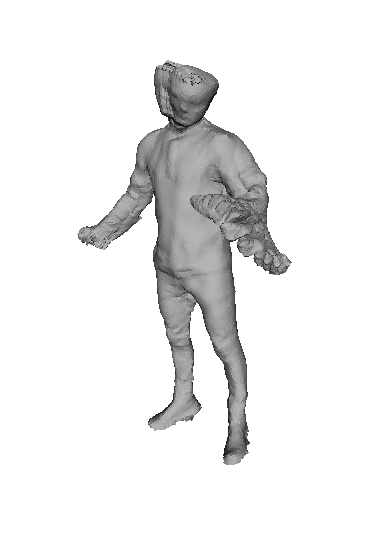}} 
	\subfigure[Golf]{ \label{fig:subfig:a} 
		\includegraphics[width=0.8in]{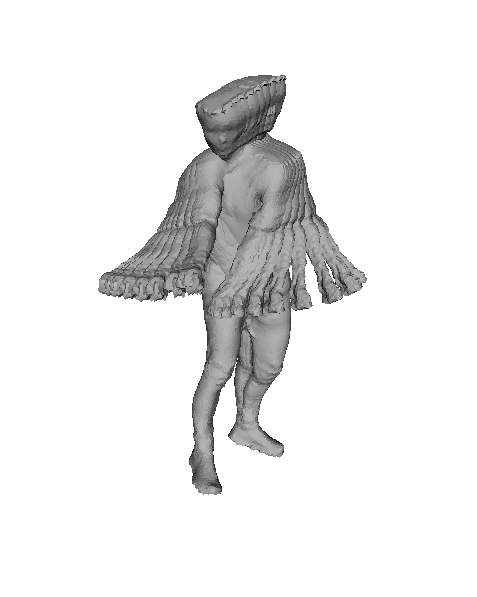}}
	\subfigure[Badminton]{ \label{fig:subfig:b} 
		\includegraphics[width=0.8in]{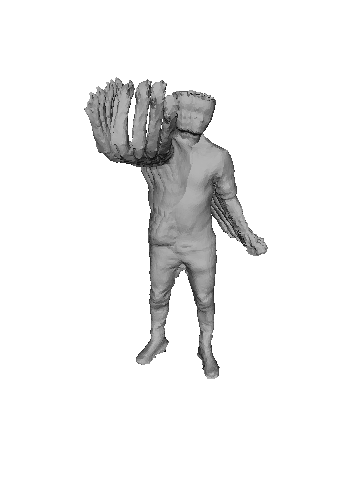}}
	\subfigure[Table tennis]{ \label{fig:subfig:c} 
		\includegraphics[width=0.8in]{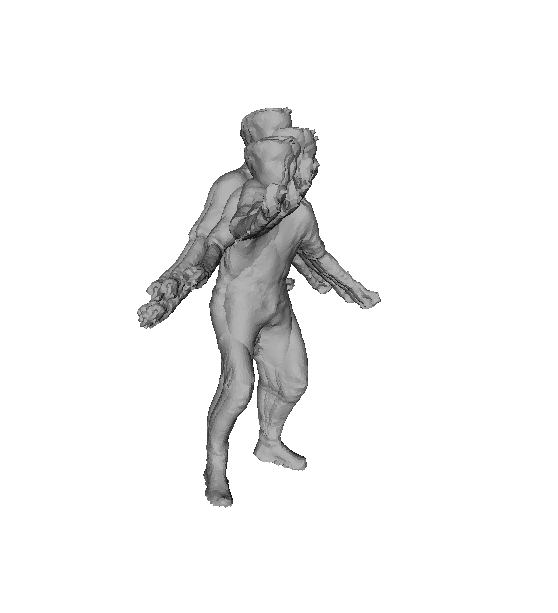}} 
	\subfigure[Weight-lifting]{ \label{fig:subfig:d} 
		\includegraphics[width=0.8in]{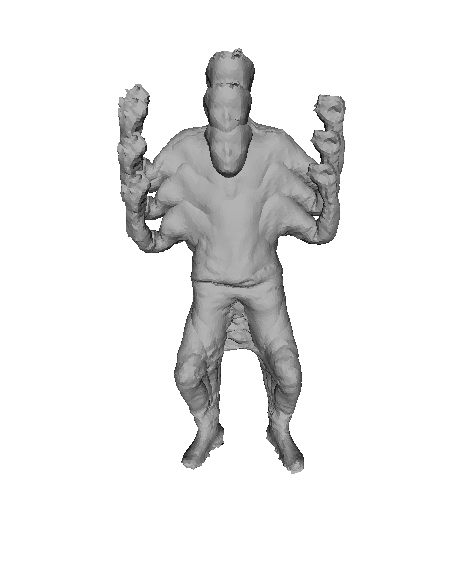}}
	\caption{Action dataset includes ten actions}\label{fig:dataset} 
\end{figure}

We constructed a 3D human action dataset using single fixed Kinect.  In order to generate a 3D whole body, we rotate the body in front of the Kinect without required of a rigid body motion, as shown in Figure  \ref{fig:reconstruction}. Then we use this model as a template to do tracking.  As shown in Figure \ref{fig:shapecompletion}  with just one view input we can get a complete body model.  Figure \ref{fig:dataset} show the sequences of reconstructed action dataset. We put a couple of frames together, so we can see the animation.  This dataset includes 10 kinds of human action class: waving, walking, bowing, clapping, kicking,  looking watch, weight-lifting, golf swinging, playing table tennis and badminton.
 

\subsection{4D-ISIP detection}

\begin{figure} 
	\centering
	\subfigure[ $\bar{H}_t$ = 0.2]{ \label{fig:subfig:a}
		\includegraphics[width=0.97in]{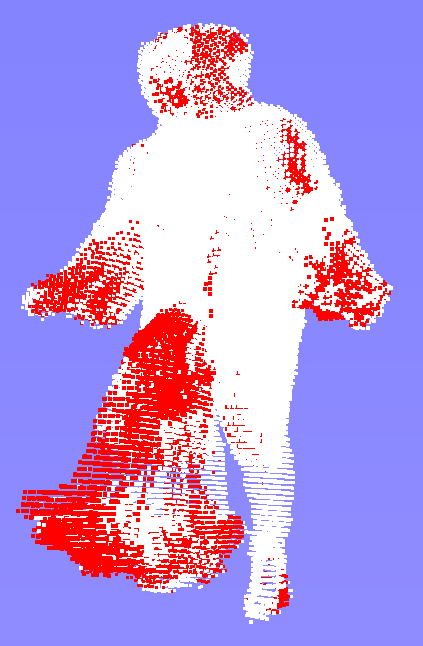}} 
	\subfigure[$\bar{H}_t$ = 0.4]{ \label{fig:subfig:b} 
		\includegraphics[width=0.95in]{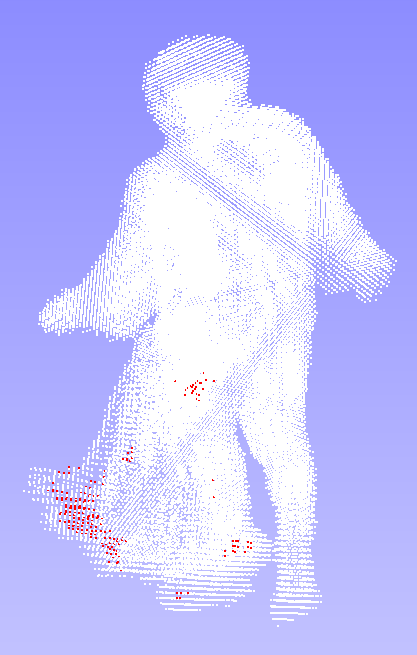}}
	\subfigure[$\bar{H}_t$ = 0.6]{ \label{fig:subfig:c} 
		\includegraphics[width=1.1in]{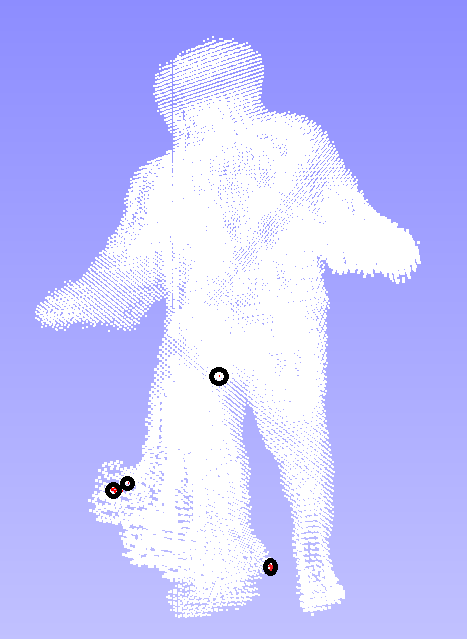}}
	\subfigure[3D sequence of kicking]{ \label{fig:subfig:d} 
		\includegraphics[width=0.95in]{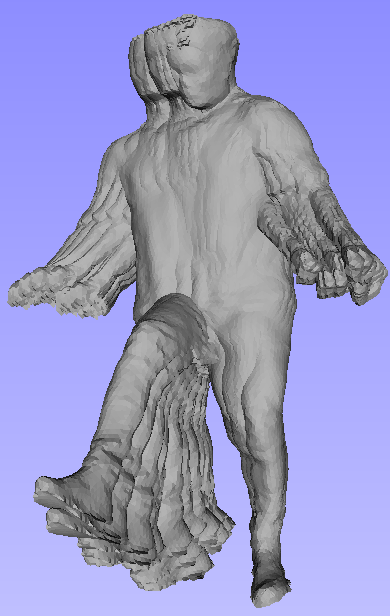}}
	\caption{Selection of threshold value $\bar{H}_t$. (a)(b)(c) are the point clouds of the action sequence, (d) is the 3D mesh of the action sequence.} 
	\label{fig:Threshold}
\end{figure}

Upon the generated human motion sequences, we can extract 4D-ISIPs. In our experiments we set the resolution of volume is 128*128*128, $\bar{\sigma}_{s} = 2$  and $\bar{\sigma}_{t} = 1$. We normalized the  value of $\bar{H}$ by $\bar{H} = (\bar{H}-\bar{H}_{min})/(\bar{H}_{max}-\bar{H}_{min})$. Apparently, we will get different results by setting different threshold. Figure \ref{fig:Threshold} (a,b,c) are the point clouds of action sequences. The red points in the those point clouds are the 4D ISIPs. As increasing the value of threshold the number of 4D-ISIPs is decreasing. Figure \ref{fig:Threshold} (d) is the 3D mesh of the kicking sequence. In order to extract sparse 4D-ISIPs, we set $\bar{H}_t= 0.6$ in following experiments.

\begin{figure}
	\centering
	\subfigure[Bowing]{	\includegraphics[width=0.7in]{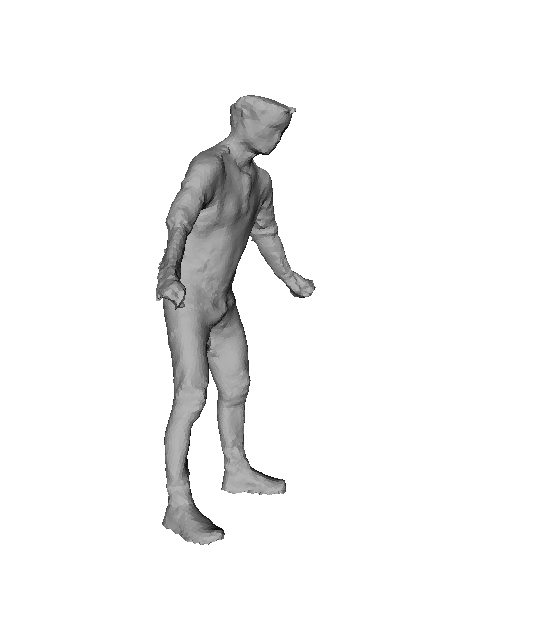}
		\includegraphics[width=0.7in]{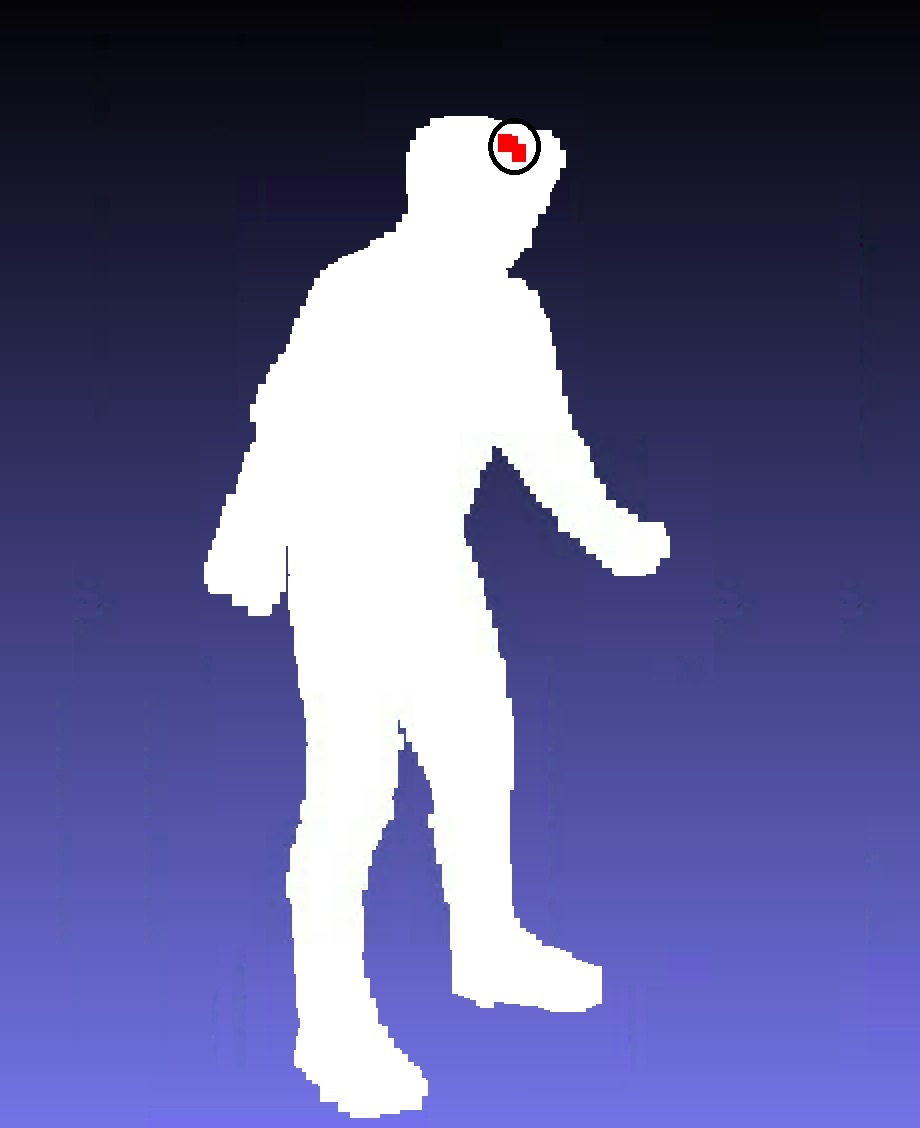}
		\includegraphics[width=0.7in]{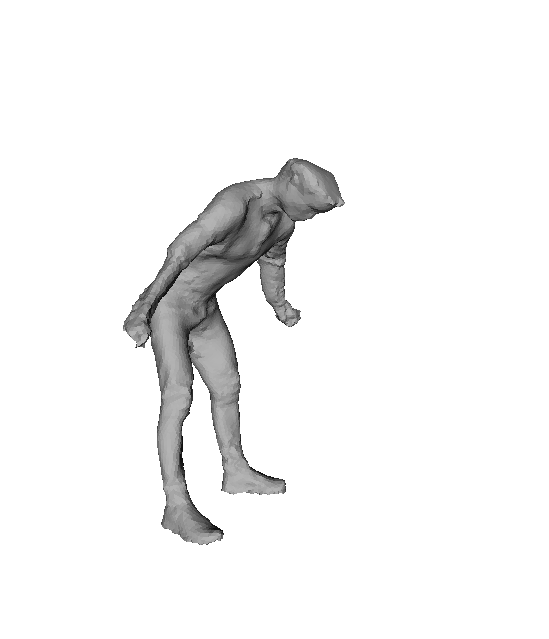}
		\includegraphics[width=0.7in]{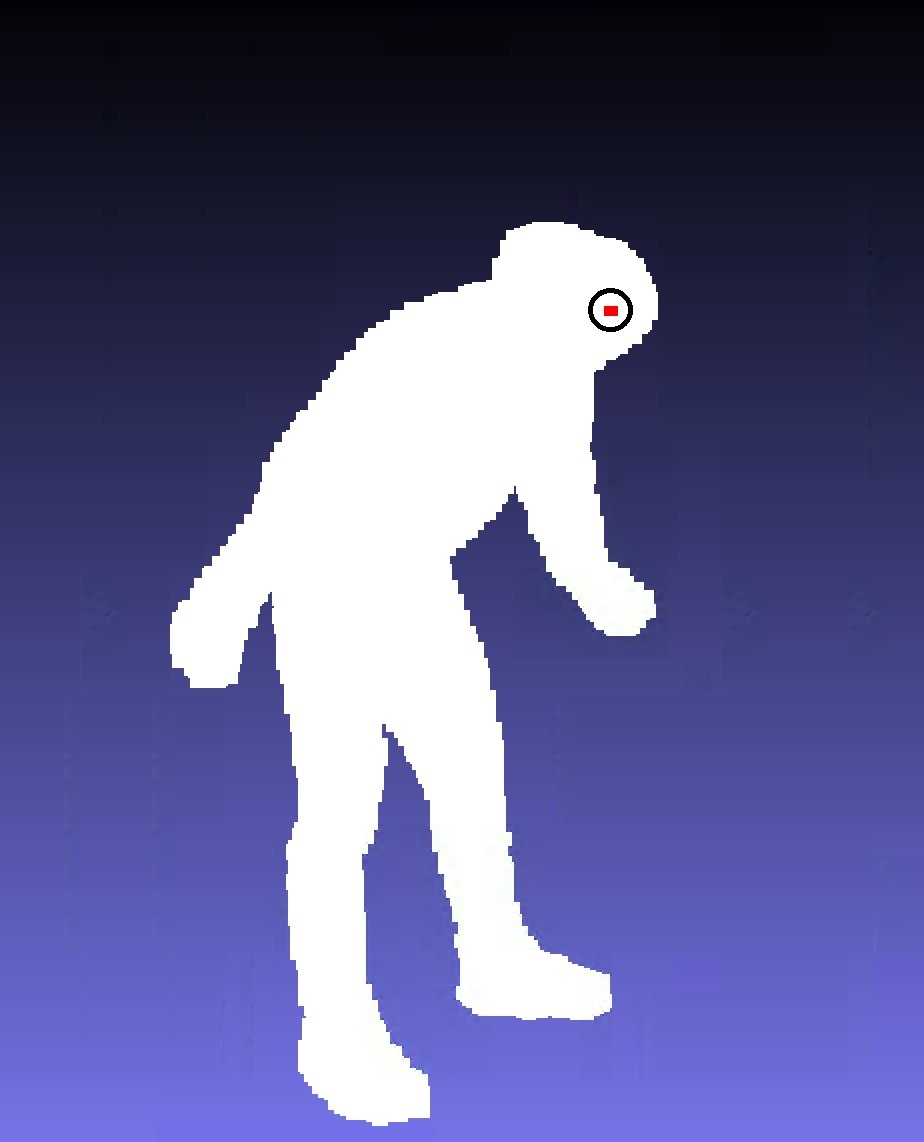}}
    \subfigure[Waving]{	\includegraphics[width=0.7in]{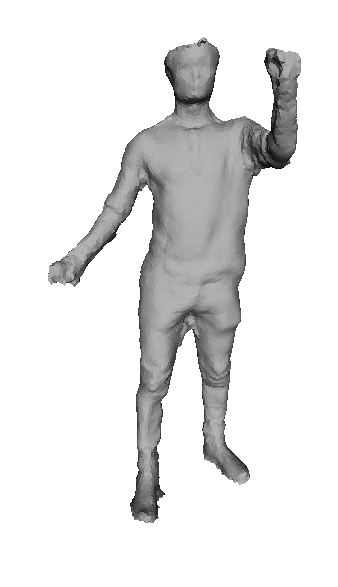}
    	\includegraphics[width=0.7in]{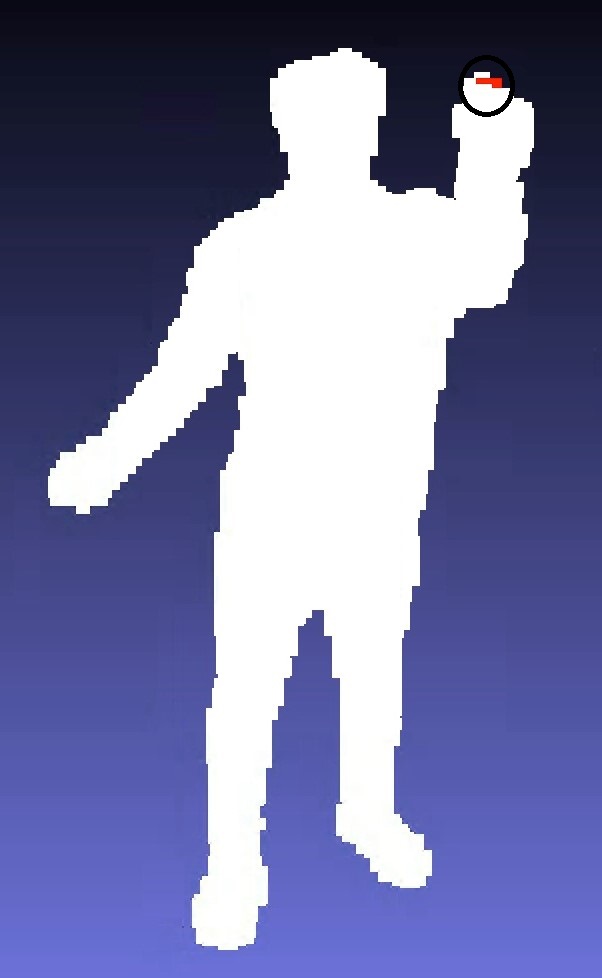} 
    	\includegraphics[width=0.7in]{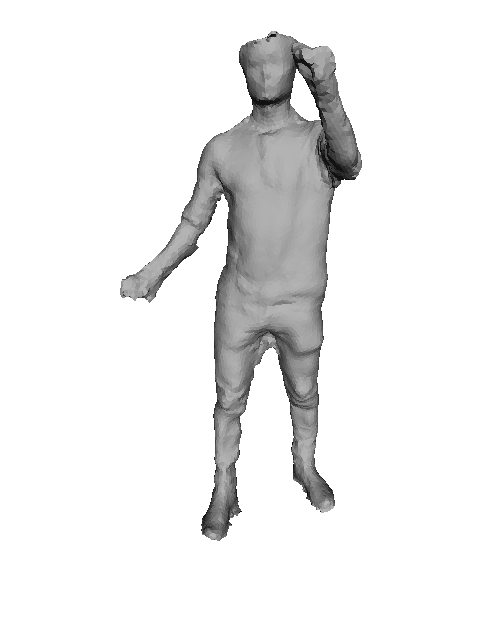}
    	\includegraphics[width=0.7in]{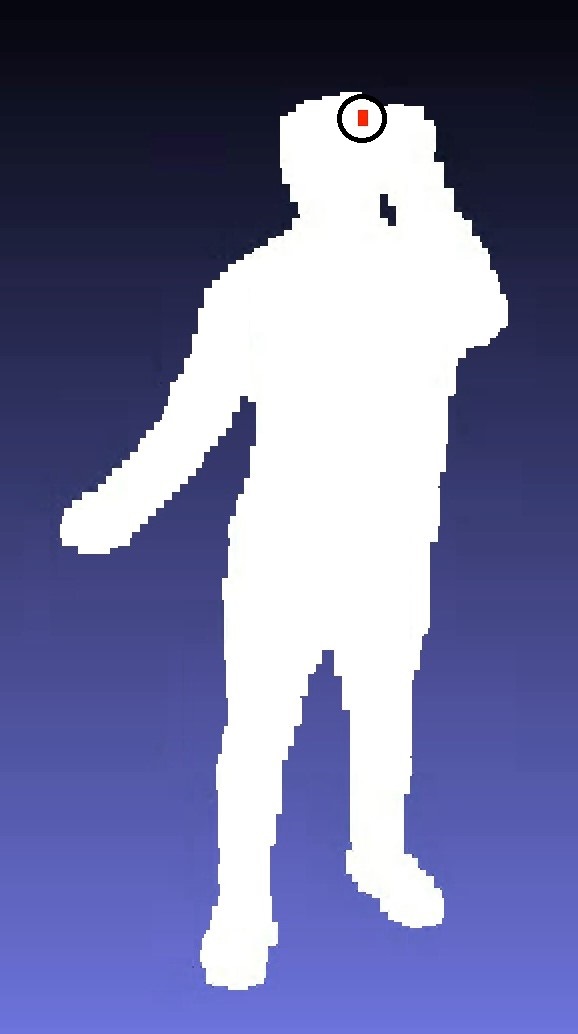}}	
    \subfigure[walking]{	
    	\includegraphics[width=0.7in]{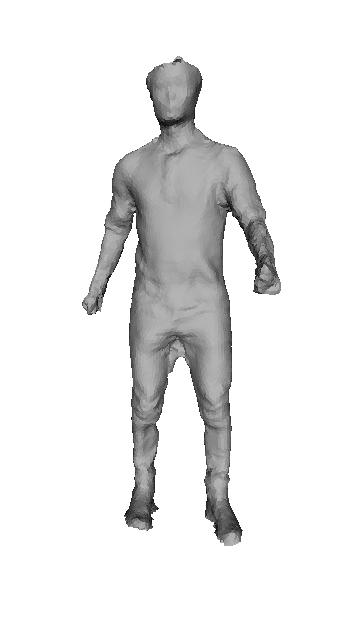}
    	\includegraphics[width=0.7in]{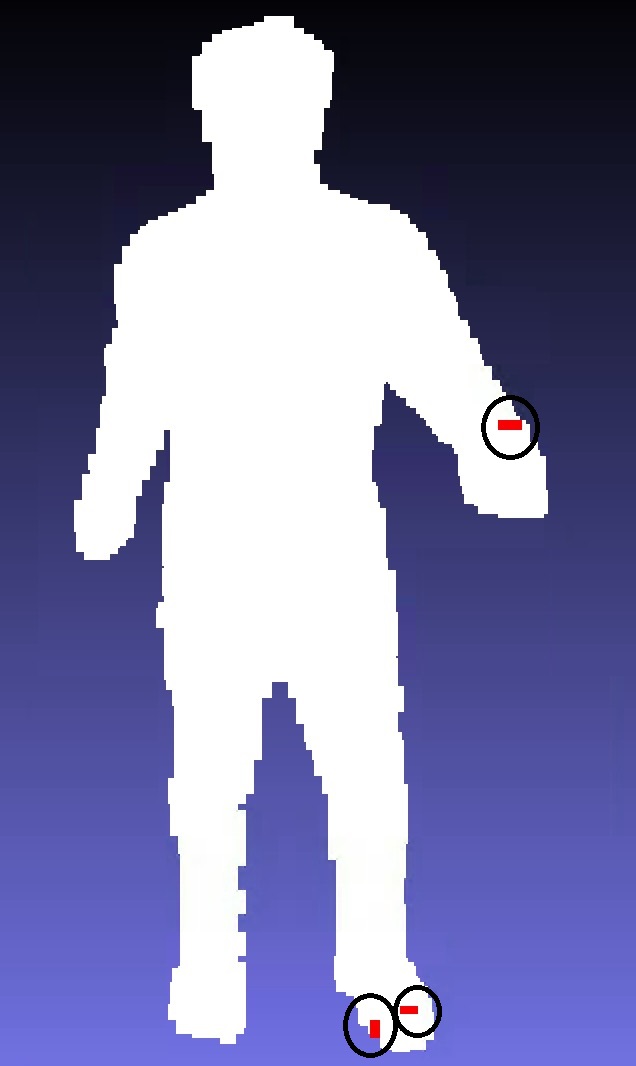}
    	\includegraphics[width=0.7in]{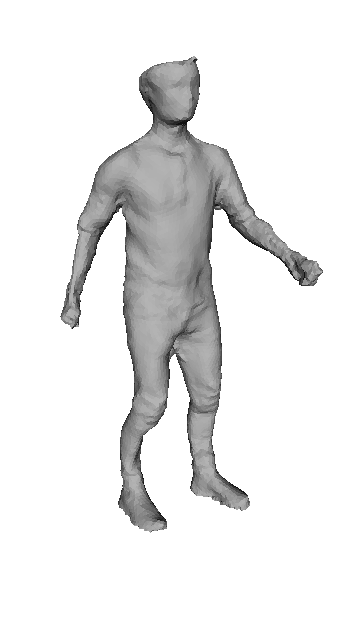}
    	\includegraphics[width=0.7in]{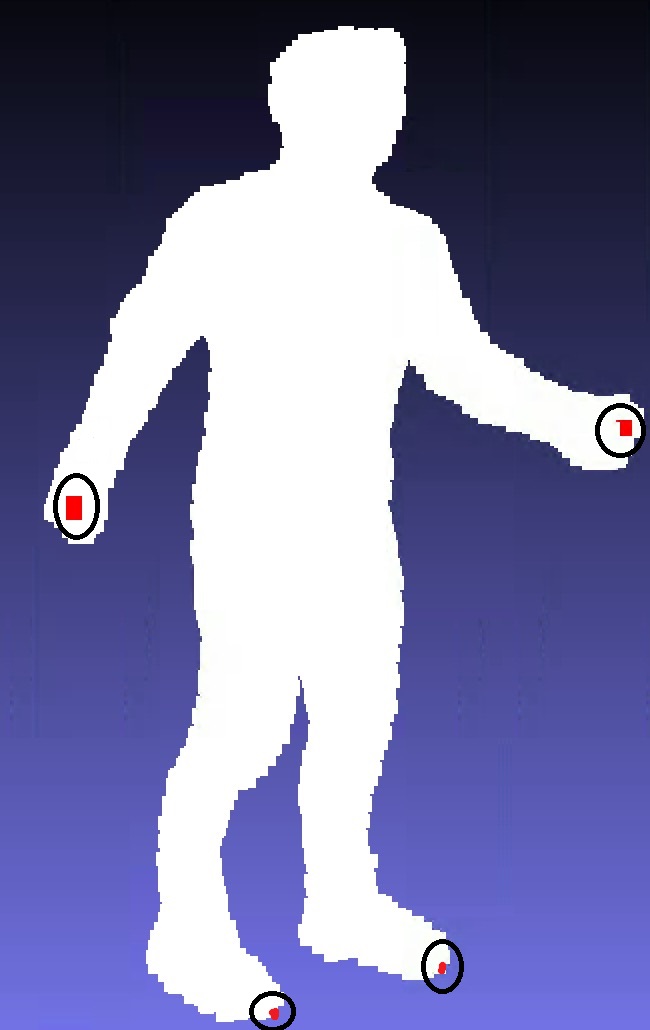}}
	\caption{4D-ISIP detection results on dataset } \label{fig:4DISIP1} 
\end{figure}

\begin{figure}
	\centering
		\subfigure[weight-lifting]{	
			\includegraphics[width=0.6in]{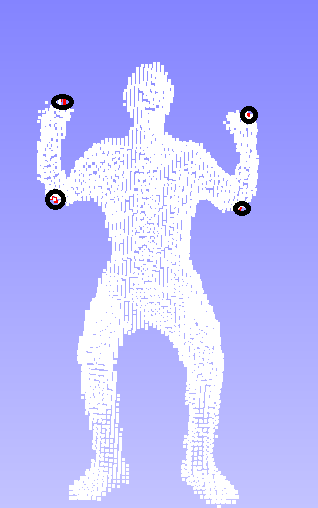}
			\includegraphics[width=0.6in]{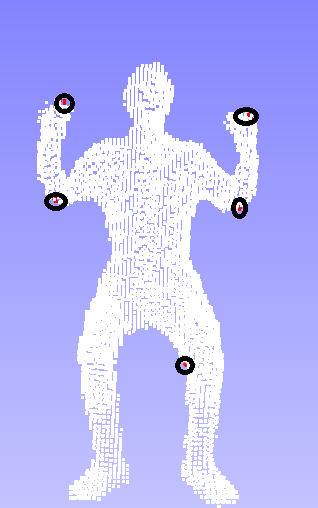}}
		\subfigure[Clapping]{	
			\includegraphics[width=0.52in]{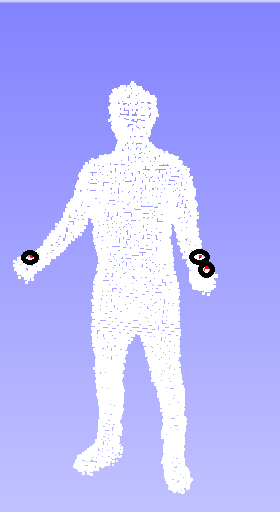}
		    \includegraphics[width=0.52in]{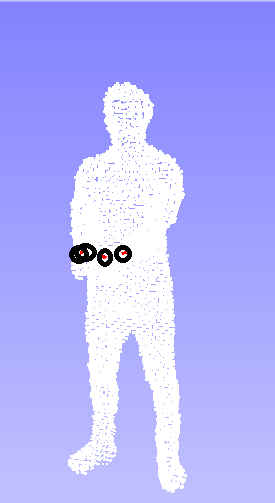}}
      \subfigure[Golf]{	
      	    \includegraphics[width=0.58in]{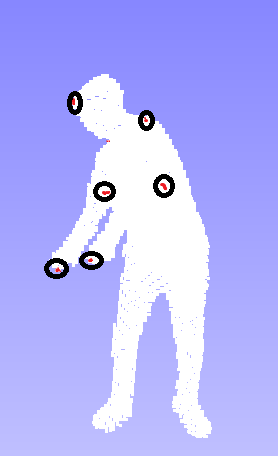}
    	    \includegraphics[width=0.49in]{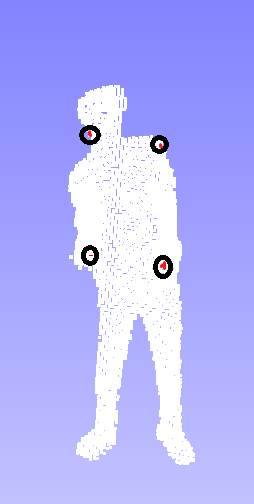}}
        
     \subfigure[Looking watch]{	
     	   \includegraphics[width=0.6in]{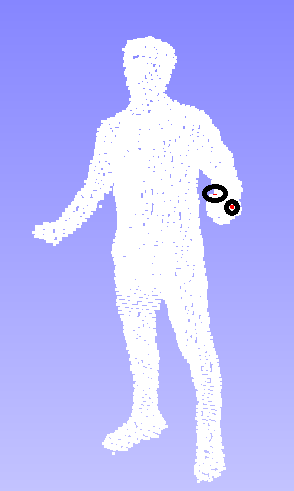}
    	   \includegraphics[width=0.56in]{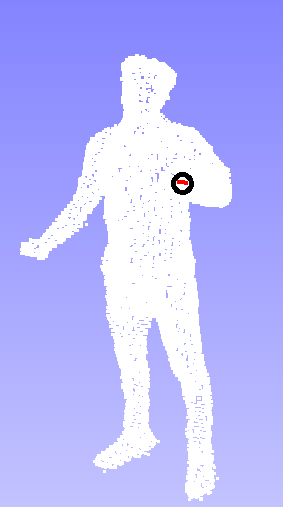}}
    \subfigure[Palying badminton]{	
    	  \includegraphics[width=0.48in]{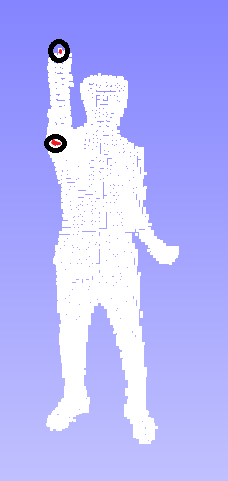}
      	  \includegraphics[width=0.66in]{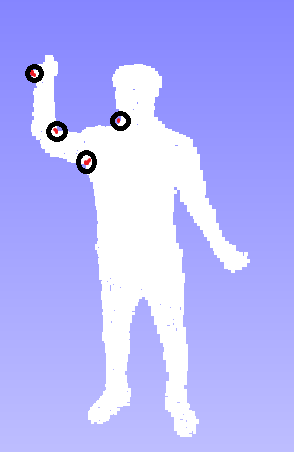}}
	\caption{More results of 4D-ISIP detection} \label{fig:4DISIP2} 
\end{figure}

As Figure \ref{fig:4DISIP1}  shows that we can robustly detect the changing motion directions, which suggests that 4D ISIP can represent the human motion. It can be used to describe trajectory of human action which can be used for action recognition.  The red points in point cloud denote the detected 4D ISIP.  The corresponding mesh models are also shown on the left. Here are more 4D ISIPs detection results in Figure \ref{fig:4DISIP2}.

\subsection{Comparison with 3D STIP}

\begin{figure} 
	\centering
		\subfigure[4D ISIP]{ \label{fig:subfig:a}
		\includegraphics[width=1.5in]{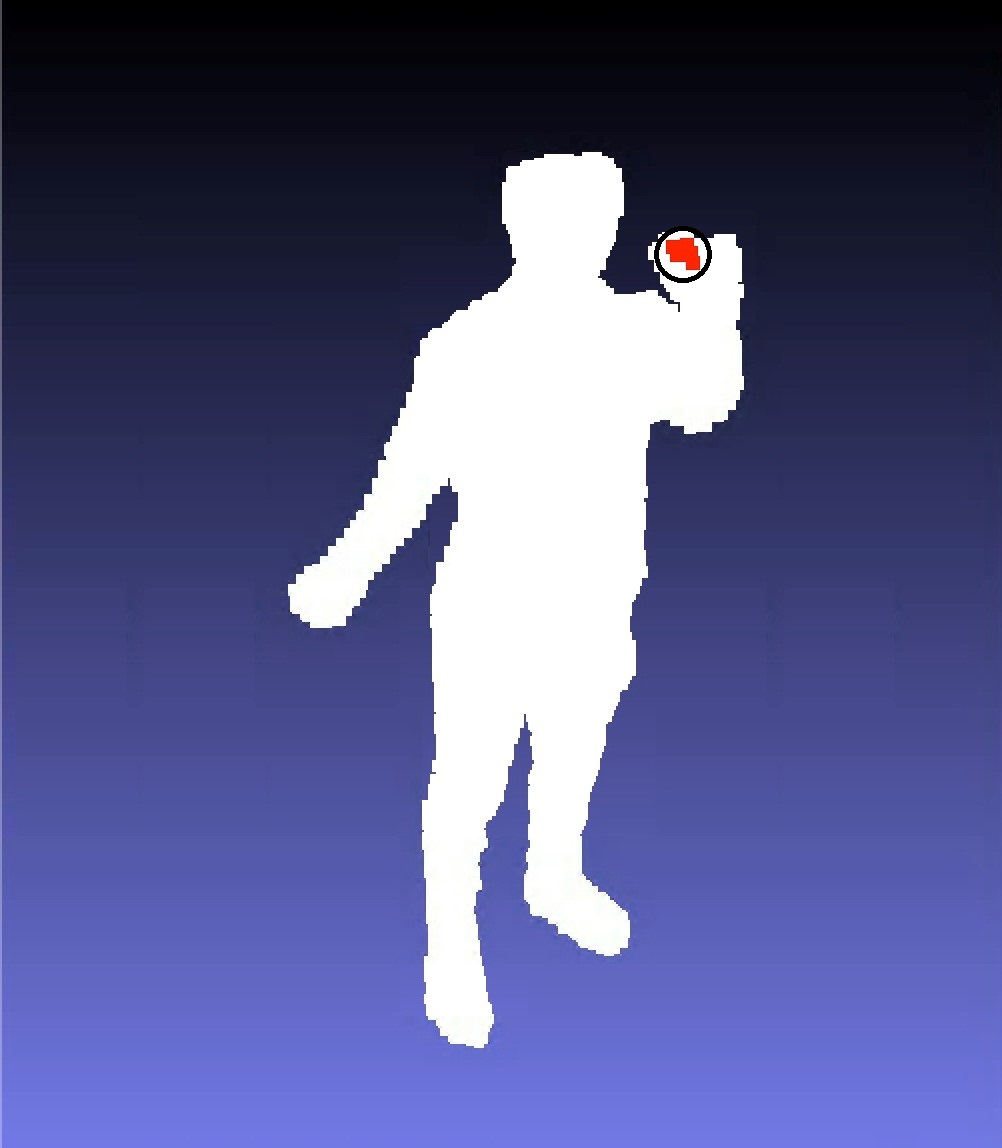}} 
		\subfigure[3D STIP ]{ \label{fig:subfig:b} 
		\includegraphics[width=1.15in]{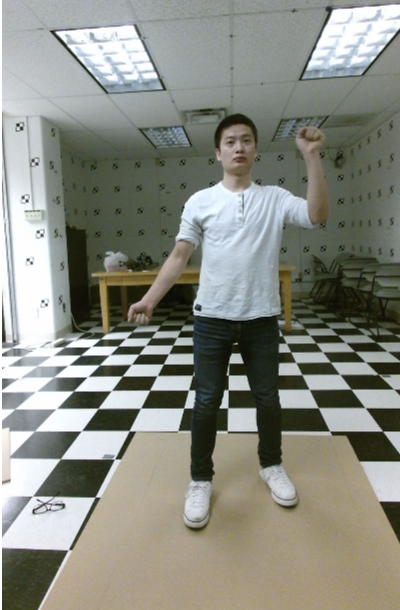}}
		\caption{4D ISIPs can be detected even there is a slightly change in image content.} 
	\label{fig:motionocclusion}
\end{figure}

Technically, 3D STIP \cite{laptev2005space} detect interest point in (x,y,t) space which can't describe the real 3D motion. It can't handle motion occlusion and illumination change problems. For instance STIP can't handle the motion of waving hand back and forth.  Because this kind of motions has slight variations in image content. However, this motion have a  significant variations in 3D (x,y,z) and 4D (x,y,z,t) space.  The proposed 4D ISIP approach can work in this situation. As shown in Figure \ref{fig:motionocclusion}. 

\begin{figure} 
	\centering
	\subfigure[3D STIP]{\includegraphics[width=0.8in]{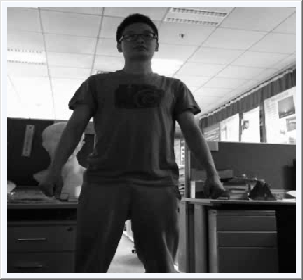} 
		\includegraphics[width=0.8in]{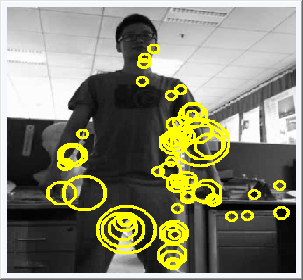}
		\includegraphics[width=0.8in]{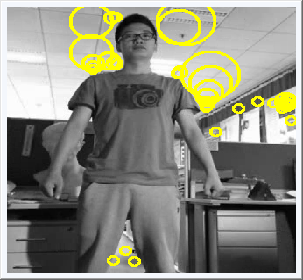}
    	\includegraphics[width=0.8in]{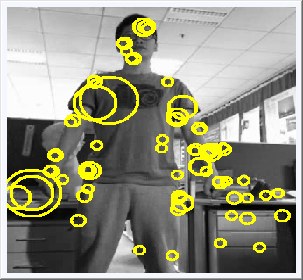}}
     \subfigure[4D ISIP]{\includegraphics[width=0.9in]{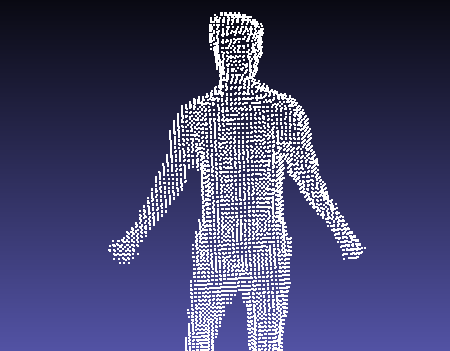} }

	\caption{Illumination change, 3D STIP  is sensitive to illumination change, (a) show the 3D STIPs on the image sequence under the illumination change. Meanwhile, 4D ISIP is robust to the illumination change, (b) is the 4D ISIP on point cloud} 
	\label{fig:lightchange}
\end{figure}

Furthermore, 4D-ISIP is robust to illumination change, which is important for action recognition. As shown in Figure \ref{fig:lightchange}. The 3D STIP is sensitive to illumination change. Because it extracts the interest points on RGB image which is sensitive to illumination change.

\section{Conclusion}
\label{sec:majhead}

In this paper, we built a system to acquire 3D human motion  using one only Kinect. We proposed a new 4D ISIP(4D implicit surface interest point detection)  which is the keypoint of the motion. It can be used for motion recognition, especially for human action recognition.We use the TSDF volume as implicit surface representation to represent the reconstructed 3D body. This results in a non-noisy human 3D action representation by fusing the depth data stream into global TSDF volume which can be useful for getting a robust interest point detection. Our approach doesn't use the image information and only uses the pure 3D geometric information to detect the interest points which has a sort of advantages. It's robust to the illumination change, occlusion, and noise caused by RGB or depth data stream. In the future work, we expected that the proposed 4D-ISIP could be used in human action recognition dealing with scale and view-invariant problems.  

\bibliographystyle{splncs03}
\bibliography{strings,refs}

\begin{thebibliography}{10}
\providecommand{\url}[1]{\texttt{#1}}
\providecommand{\urlprefix}{URL }

\bibitem{bogo2015detailed}
Bogo, F., Black, M.J., Loper, M., Romero, J.: Detailed full-body
  reconstructions of moving people from monocular rgb-d sequences. In:
  Proceedings of the IEEE International Conference on Computer Vision. pp.
  2300--2308 (2015)

\bibitem{chen2013tensor}
Chen, Y., Liu, Z., Zhang, Z.: Tensor-based human body modeling. In: Proceedings
  of the IEEE Conference on Computer Vision and Pattern Recognition. pp.
  105--112 (2013)

\bibitem{cho2015volumetric}
Cho, S.S., Lee, A.R., Suk, H.I., Park, J.S., Lee, S.W.: Volumetric spatial
  feature representation for view-invariant human action recognition using a
  depth camera. Optical Engineering  54(3),  033102--033102 (2015)

\bibitem{curless1996volumetric}
Curless, B., Levoy, M.: A volumetric method for building complex models from
  range images. In: Proceedings of the 23rd annual conference on Computer
  graphics and interactive techniques. pp. 303--312. ACM (1996)

\bibitem{dollar2005behavior}
Doll{\'a}r, P., Rabaud, V., Cottrell, G., Belongie, S.: Behavior recognition
  via sparse spatio-temporal features. In: Visual Surveillance and Performance
  Evaluation of Tracking and Surveillance, 2005. 2nd Joint IEEE International
  Workshop on. pp. 65--72. IEEE (2005)

\bibitem{gkalelis2009i3dpost}
Gkalelis, N., Kim, H., Hilton, A., Nikolaidis, N., Pitas, I.: The i3dpost
  multi-view and 3d human action/interaction database. In: Visual Media
  Production, 2009. CVMP'09. Conference for. pp. 159--168. IEEE (2009)

\bibitem{guo2015robust}
Guo, K., Xu, F., Wang, Y., Liu, Y., Dai, Q.: Robust non-rigid motion tracking
  and surface reconstruction using l0 regularization. In: Proceedings of the
  IEEE International Conference on Computer Vision. pp. 3083--3091 (2015)

\bibitem{harris1988combined}
Harris, C., Stephens, M.: A combined corner and edge detector. In: Alvey vision
  conference. vol.~15, p.~50. Citeseer (1988)

\bibitem{holte2012local}
Holte, M.B., Chakraborty, B., Gonzalez, J., Moeslund, T.B.: A local 3-d motion
  descriptor for multi-view human action recognition from 4-d spatio-temporal
  interest points. Selected Topics in Signal Processing, IEEE Journal of  6(5),
   553--565 (2012)

\bibitem{kim2014view}
Kim, S.J., Kim, S.W., Sandhan, T., Choi, J.Y.: View invariant action
  recognition using generalized 4d features. Pattern Recognition Letters  49,
  40--47 (2014)

\bibitem{kuehne2011hmdb}
Kuehne, H., Jhuang, H., Garrote, E., Poggio, T., Serre, T.: Hmdb: a large video
  database for human motion recognition. In: Computer Vision (ICCV), 2011 IEEE
  International Conference on. pp. 2556--2563. IEEE (2011)

\bibitem{laptev2005space}
Laptev, I.: On space-time interest points. International Journal of Computer
  Vision  64(2-3),  107--123 (2005)

\bibitem{li2010action}
Li, W., Zhang, Z., Liu, Z.: Action recognition based on a bag of 3d points. In:
  Computer Vision and Pattern Recognition Workshops (CVPRW), 2010 IEEE Computer
  Society Conference on. pp. 9--14. IEEE (2010)

\bibitem{lowe2004distinctive}
Lowe, D.G.: Distinctive image features from scale-invariant keypoints.
  International journal of computer vision  60(2),  91--110 (2004)

\bibitem{newcombe2015dynamicfusion}
Newcombe, R.A., Fox, D., Seitz, S.M.: Dynamicfusion: Reconstruction and
  tracking of non-rigid scenes in real-time. In: Proceedings of the IEEE
  Conference on Computer Vision and Pattern Recognition. pp. 343--352 (2015)

\bibitem{niebles2010modeling}
Niebles, J.C., Chen, C.W., Fei-Fei, L.: Modeling temporal structure of
  decomposable motion segments for activity classification. In: European
  conference on computer vision. pp. 392--405. Springer (2010)

\bibitem{reddy2013recognizing}
Reddy, K.K., Shah, M.: Recognizing 50 human action categories of web videos.
  Machine Vision and Applications  24(5),  971--981 (2013)

\bibitem{rodriguez2008action}
Rodriguez, M.D., Ahmed, J., Shah, M.: Action mach a spatio-temporal maximum
  average correlation height filter for action recognition. In: Computer Vision
  and Pattern Recognition, 2008. CVPR 2008. IEEE Conference on. pp. 1--8. IEEE
  (2008)

\bibitem{rosten2010faster}
Rosten, E., Porter, R., Drummond, T.: Faster and better: A machine learning
  approach to corner detection. Pattern Analysis and Machine Intelligence, IEEE
  Transactions on  32(1),  105--119 (2010)

\bibitem{shi1994good}
Shi, J., Tomasi, C.: Good features to track. In: Computer Vision and Pattern
  Recognition, 1994. Proceedings CVPR'94., 1994 IEEE Computer Society
  Conference on. pp. 593--600. IEEE (1994)

\bibitem{tran2008human}
Tran, D., Sorokin, A.: Human activity recognition with metric learning.
  Computer Vision--ECCV 2008 pp. 548--561 (2008)

\bibitem{wang2009evaluation}
Wang, H., Ullah, M.M., Klaser, A., Laptev, I., Schmid, C.: Evaluation of local
  spatio-temporal features for action recognition. In: BMVC 2009-British
  Machine Vision Conference. pp. 124--1. BMVA Press (2009)

\bibitem{wang2012mining}
Wang, J., Liu, Z., Wu, Y., Yuan, J.: Mining actionlet ensemble for action
  recognition with depth cameras. In: Computer Vision and Pattern Recognition
  (CVPR), 2012 IEEE Conference on. pp. 1290--1297. IEEE (2012)

\bibitem{weinland2007action}
Weinland, D., Boyer, E., Ronfard, R.: Action recognition from arbitrary views
  using 3d exemplars. In: Computer Vision, 2007. ICCV 2007. IEEE 11th
  International Conference on. pp. 1--7. IEEE (2007)

\bibitem{weiss2011home}
Weiss, A., Hirshberg, D., Black, M.J.: Home 3d body scans from noisy image and
  range data. In: Computer Vision (ICCV), 2011 IEEE International Conference
  on. pp. 1951--1958. IEEE (2011)

\bibitem{willems2008efficient}
Willems, G., Tuytelaars, T., Van~Gool, L.: An efficient dense and
  scale-invariant spatio-temporal interest point detector. In: Computer
  Vision--ECCV 2008, pp. 650--663. Springer (2008)

\bibitem{xia2013spatio}
Xia, L., Aggarwal, J.: Spatio-temporal depth cuboid similarity feature for
  activity recognition using depth camera. In: Proceedings of the IEEE
  Conference on Computer Vision and Pattern Recognition. pp. 2834--2841 (2013)

\bibitem{zhang20114}
Zhang, H., Parker, L.E.: 4-dimensional local spatio-temporal features for human
  activity recognition. In: Intelligent robots and systems (IROS), 2011
  IEEE/RSJ international conference on. pp. 2044--2049. IEEE (2011)

\bibitem{zhang2014quality}
Zhang, Q., Fu, B., Ye, M., Yang, R.: Quality dynamic human body modeling using
  a single low-cost depth camera. In: Computer Vision and Pattern Recognition
  (CVPR), 2014 IEEE Conference on. pp. 676--683. IEEE (2014)

\bibitem{zhu2014evaluating}
Zhu, Y., Chen, W., Guo, G.: Evaluating spatiotemporal interest point features
  for depth-based action recognition. Image and Vision Computing  32(8),
  453--464 (2014)

\end{thebibliography}

\end{document}